\documentclass[letterpaper, 10 pt, conference]{ieeeconf}

\IEEEoverridecommandlockouts                              

\usepackage{flushend}
\usepackage{hyperref}
\usepackage{graphics} 
\usepackage{epsfig} 
\usepackage{mathptmx} 
\usepackage{times} 
\usepackage{amsmath} 
\usepackage{amssymb}  
\usepackage{xcolor}

\DeclareMathOperator*{\argmax}{arg\,max}
\DeclareMathOperator*{\argmin}{arg\,min}
\usepackage{tabularx}

\usepackage{amsfonts}
\usepackage{bm}
\usepackage{bbm}
\usepackage{cite}

\title{\LARGE \bf
Scaled Autonomy: Enabling Human Operators to Control Robot Fleets
}

\author{Gokul Swamy$^{1}$, Siddharth Reddy$^{1}$, Sergey Levine$^{1}$, Anca D. Dragan$^{1}$
\thanks{$^{1}$Department of Electrical Engineering and Computer Science,
  University of California, Berkeley}%
}

\begin{document}

\maketitle
\thispagestyle{empty}
\pagestyle{empty}

\begin{abstract}
Autonomous robots often encounter challenging situations where their control policies fail and an expert human operator must briefly intervene, e.g., through teleoperation.
In settings where multiple robots act in separate environments,
a single human operator can manage a fleet of robots by identifying and teleoperating one robot at any given time.
The key challenge is that users have limited attention: as the number of robots increases, users lose the ability to decide which robot requires teleoperation the most.
Our goal is to automate this decision, thereby enabling users to supervise more robots than their attention would normally allow for.
Our insight is that we can model the user's choice of which robot to control as an approximately optimal decision that maximizes the user's utility function.
We learn a model of the user's preferences from observations of the user's choices in easy settings with a few robots, and use it in challenging settings with more robots to automatically identify which robot the user would most likely choose to control, if they were able to evaluate the states of all robots at all times.
We run simulation experiments and a user study with twelve participants that show our method can be used to assist users in performing a simulated navigation task.
We also run a hardware demonstration that illustrates how our method can be applied to a real-world mobile robot navigation task.
\end{abstract}

\section{Introduction}

Sliding autonomy \cite{dias2008sliding,kortenkamp2000adjustable,bruemmer2002dynamic,sellner2005user} is a promising approach to deploying robots with imperfect control policies: while a fleet of autonomous robots acts in separate environments, a human operator monitors their states, and can intervene to help a robot via teleoperation when the robot encounters a challenging state.
Imagine, for instance, a fleet of delivery robots: at any given time, most of them may be driving in easy conditions with few pedestrians, while one of them encounters a crowded sidewalk and requires operator intervention.
Ideally, the performance of such a human-robot centaur team would scale smoothly with the increasing capabilities of the robot: as autonomy improves, challenging states become rarer, and a single human operator should be able to control a larger fleet of robots.

Unfortunately, while the user may be a skilled operator, they have limited attention. As the fleet grows larger, the user's ability to maintain awareness of the states of all robots and identify the robot that most requires intervention degrades.

\begin{figure}
  \includegraphics[width=\linewidth]{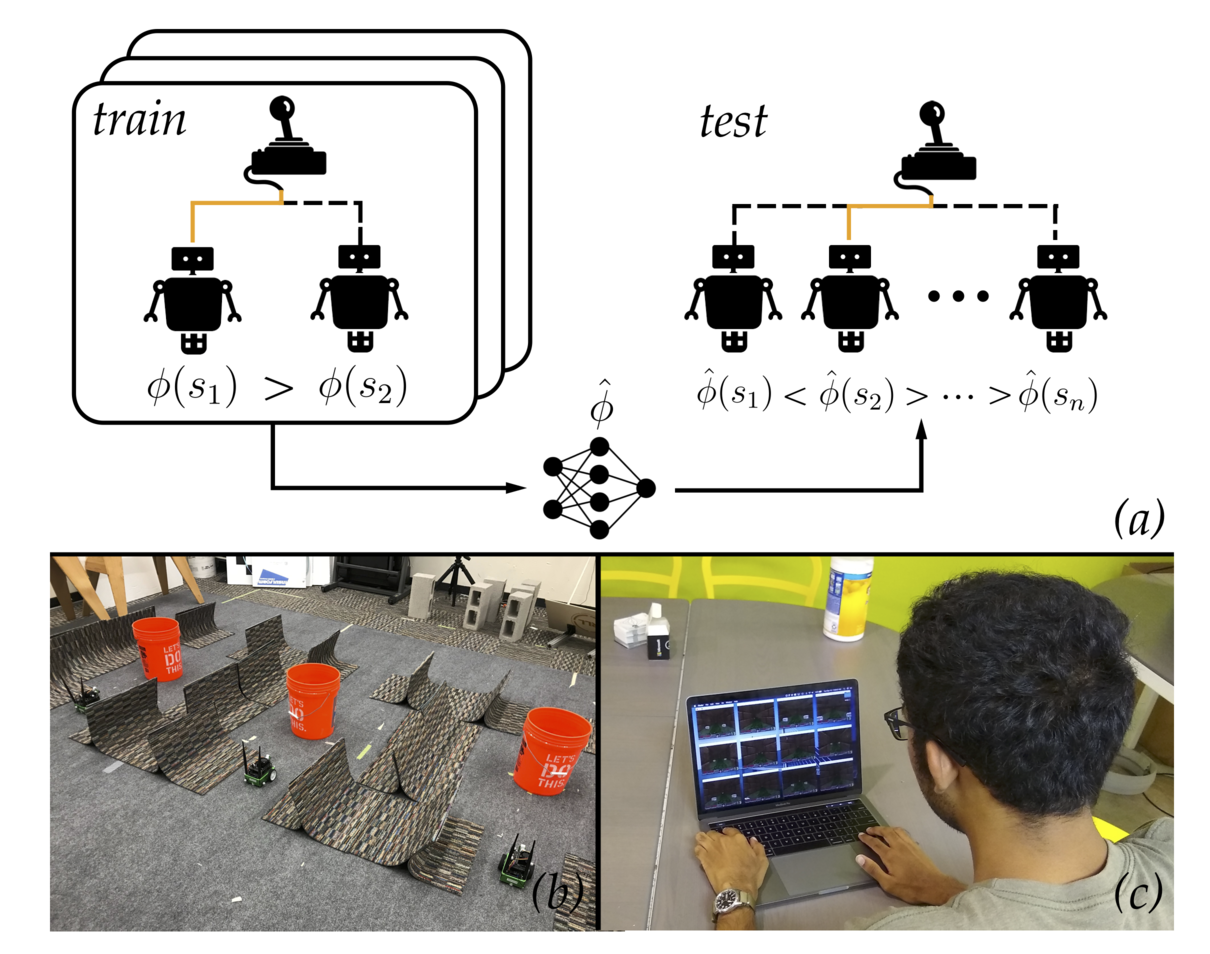}
  \caption{We learn which robot a user would prefer to control, by observing the user manage a small number of robots. We then use the learned preference model to help the user control a much larger number of robots. We evaluate our method (a) on a simulated navigation task and real hardware demonstration (b) through controlled, synthetic experiments with expert agents that stand in for users, and a human user study with twelve participants (c).}
  \label{fig:front}
\end{figure}

We propose to overcome this challenge by automating the operator's choice of which robot to control.
Given a large number of robots running in separate environments, our approach is to train a model that predicts which robot the user would teleoperate, if the user had the ability to analyze all the robots' current states quickly enough.
Our insight is that we can use decisions that the operator makes in easy settings with only a few robots, where they can feasibly pay attention to all the robots' current states, to train a predictive model of user behavior that generalizes to challenging settings with many robots.

The key to generalizing the user's choices from easy to hard settings is to treat them as observations of relative preferences between robots: every choice the person makes to control one particular robot instead of any of the other robots is assumed to be an approximately optimal decision, with respect to maximizing the user's utility function.
Every choice gives us information about that utility, namely that the utility of controlling the chosen robot was higher than the utility of controlling any other robot.
We can thus use observations of the operator's choices to fit a model of their utility function.
At test time, we apply the learned model to the current state of each robot, and automatically switch the user to controlling the robot with the highest predicted likelihood of being chosen.

We test our method in simulation and through an in-person user study, on a simulated navigation task and real hardware demonstration.
In the navigation task, the robot must successfully navigate through a video game environment with hazards and health packs to reach a goal state (see schematic in Figure \ref{fig:study_perf}).
We initially evaluate our method in synthetic experiments where we simulate user input under ideal assumptions, and where we have access to ground-truth user preferences.
We find that our method effectively generalizes the user's choices in easy settings with a small number of robots to challenging settings with a large number of robots.
We also find that modeling the user's choices as a function of relative preferences between robots is important for this generalization.
To show that our results extend to real user data, we conduct an in-person user study with twelve human participants, where we evaluate each participant's ability to manage twelve robots with and without assisted choice.
We find that assisted choice enables users to perform significantly better than they can on their own.

\section{Related Work}

In shared autonomy, a human operator and robot collaborate to control a system which neither the operator nor the robot could control effectively by themselves \cite{aigner1997human}.
Previous work in this area \cite{javdani2015shared,dragan2013policy,reddy2018shared,schwarting2017parallel,broad2019highly}
has focused on some combination of inferring user intent and acting to achieve it.
We instead focus on helping the user process information quickly enough to manage a fleet of robots.
The problem we tackle is more akin to that addressed by a continuously-running search engine like the Remembrance Agent \cite{rhodes1996remembrance}, which assists a user's decision-making by displaying information relevant to the user's current context.

The closest prior work is in sliding autonomy \cite{dias2008sliding,kortenkamp2000adjustable,bruemmer2002dynamic,sellner2005user} and active learning \cite{cohn2011comparing}, where the robot can request user intervention in challenging or uncertain situations.
Prior methods tend to require knowledge of the task in order to determine when user intervention is needed.
Our method makes minimal assumptions about the task, and instead allocates the user's attention using a learned model of the user's preferences.
To our knowledge, we are the first to use a general-purpose learning approach to allocating operator interventions.

\section{Learning to Allocate\\Operator Interventions}

Our goal is to help the user choose which robot to control at any given time.
To do so, we learn to mimic the way the user manages a small number of robots, then use the learned model to assist the user in controlling a large number of robots.

\subsection{Problem Formulation} \label{problem-formulation}

We formalize the problem of automating the operator's decision of which robot to control as one of estimating the operator's \emph{internal scoring function}: a function that maps the state of a robot to a real-valued score of how useful it would be to take control of the robot, in terms of maximizing cumulative task performance across robots.

\noindent\textbf{User choice model.}
Let $[n]$ denote $\{1, 2, ..., n\}$, $i \in [n]$ denote the $i$-th robot in a fleet of $n$ robots, and $i_H^t \in [n]$ denote the robot controlled by the user at time $t$.
We assume the user selects robot $i_H^t$ using the Luce choice model \cite{luce2012individual},
\begin{equation} \label{eq:luce}
\mathbb{P}[i_H^t = i] = e^{\phi(s_i^t)} / \sum_{j=1}^n e^{\phi(s_j^t)},
\end{equation}
where $\phi : \mathcal{S} \to \mathbb{R}$ is the user's scoring function, and $s_i^t$ is the state of robot $i$ at time $t$.
In other words, we assume users choose to control higher-scoring robots with exponentially higher probability.
Crucially, we also assume that the score of each robot is independent of the other robots.
This makes it possible to scale the model to a large number of robots $n$, which would not be practical if, e.g., scores depended on interactions between the states of different robots.

\noindent\textbf{User rationality model.}
We assume the user's control policy $\pi_H : \mathcal{S} \times \mathcal{A} \to [0, 1]$ maximizes the user's utility function $R$.
At time $t$, the user chooses a robot $i_H^t$ to control, then controls it using their policy $\pi_H$.
We assume the user's scoring function $\phi$ maximizes the cumulative task performance of all robots:
\begin{equation} \label{eq:cumulative-reward-objective}
\phi = \argmax_{\phi \in \Phi} \mathbb{E}\left[\sum_{i=1}^n \sum_{t=0}^{T-1} R(s_i^t, a_i^t) \mid \pi_H, \pi_R\right],
\end{equation}
where $T$ is the episode horizon.
The actions $a_i^t$ are determined by whether the user was in control and executed their policy $\pi_H$, or the robot relied on its own policy $\pi_R$. Formally,
\begin{equation}
\mathbb{P}[a_i^t = a] = \mathbb{P}[i_H^t = i]\pi_H(a | s_i^t) + (1 - \mathbb{P}[i_H^t = i])\pi_R(a | s_i^t),
\end{equation}
where the user's choice $\mathbb{P}[i_H^t = i]$ of whether or not to control robot $i$ is modeled in Equation \ref{eq:luce}.\footnote{We assume the user's scoring function $\phi$ is optimal with respect to the utility objective in Equation \ref{eq:cumulative-reward-objective} for the sake of clarity. However, our method is still useful in settings where $\phi$ is suboptimal. In such settings, our method will, at best, match the performance of the user's suboptimal $\phi$.}

\noindent\textbf{Robot policy.}
We assume the robot policy $\pi_R$ is identical for each of the $n$ robots, and that it does not perfectly maximize the user's utility function $R$.
Our method is agnostic to how $\pi_R$ is constructed: e.g., it could be a decision tree of hard-coded control heuristics, or a planning algorithm equipped with a forward dynamics model of the environment.
In this work, we choose $\pi_R$ to be a learned policy that is trained to imitate user actions.
This allows us to make minimal assumptions about the task and environment, and enables the robot policy to improve as the amount and quality of user demonstration data increases.

\noindent\textbf{Knowns and unknowns.}
We assume we know the robot policy $\pi_R$, but we do not know the user's utility function $R$, the user's control policy $\pi_H$, or the user's scoring function $\phi$.\footnote{We assume the user's utility function $R$ is unknown for the sake of clarity. However, our method is also useful in settings where the user's utility function $R$ is known, but the utility-maximizing policy $\pi_H$ is difficult to compute or learn from demonstrations.}

\noindent\textbf{Problem statement.}
Our assumptions about the user's rationality may not hold in settings with a large number of robots $n$: because the user has limited attention, they may not be able to evaluate the scores of the states of all robots at all times using their internal scoring function $\phi$.
As a result, they may make systematically suboptimal choices that do not maximize the expected cumulative task performance across all robots.

\subsection{Our Method} \label{our-method}

Our aim is to help the user maximize the expected cumulative task performance across all robots, in settings with a large number of robots $n$.
We do so by learning a scoring model $\hat{\phi}$ and using it to automate the user's choice of which robot to control.
In conjunction, we train the robot policy $\pi_R$ to imitate user action demonstrations -- collected initially on a single robot, then augmented with additional demonstrations collected as the user operates each robot chosen by the scoring model $\hat{\phi}$.

We split our method into four phases.
In phase one, we train the robot policy $\pi_R$ using imitation learning.
The user controls a single robot in isolation, and we collect a demonstration dataset $\mathcal{D}_{\text{demo}}$ of state-action pairs $(s, a)$.
Our method is agnostic to the choice of imitation learning algorithm.
In phase two, we learn a scoring model $\hat{\phi}$ by asking the user to manage a small number of robots, observing which robot $i_H^t$ the user chooses to control at each timestep based on their internal scoring function $\phi$, then fitting a parametric model of $\phi$ that explains the observed choices.
In phase three, we enable the user to manage a large number of robots by using the learned scoring model $\hat{\phi}$ to automatically choose which robot to control for them.
In phase four, we update the robot's imitation policy $\pi_R$ with the newly-acquired user action demonstrations from phase three.

\noindent\textbf{Phase one (optional): training the robot policy $\pi_R$.}
In this work, we train the robot policy $\pi_R$ to imitate the user policy $\pi_H$.
We record state-action demonstrations $\mathcal{D}_{\text{demo}}$ generated by the user as they control one robot in isolation, and use those demonstrations to train a policy that each robot can execute autonomously during phases two and three.
We implement $\pi_R$ using a simple nearest-neighbor classifier that selects the action taken in the closest state for which we have a demonstration from $\pi_H$.
Formally,
\begin{equation}
\pi_R(a | s) =
  \begin{cases}
    1 & \text{ if } (\cdot, a) = \argmin_{(s', a') \in \mathcal{D}_{\text{demo}}} \|s - s'\|_2 \\
    0 & \text{ otherwise}. \\
  \end{cases}
\end{equation}
We choose a simple imitation policy for $\pi_R$ in order to model real-world tasks for which even state-of-the-art robot control policies are suboptimal.
Improving the autonomous robot policy $\pi_R$ is orthogonal to the objective of this paper, which is to enable an arbitrary robot policy $\pi_R$ to be improved by the presence of a human operator capable of intervening in challenging states.

\noindent\textbf{Phase two: learning the scoring model $\hat{\phi}$.}
Our approach to assisting the user involves estimating their scoring function $\phi$.
To do so, we have the user manage a small number of robots $n$.
While the user operates one robot using control policy $\pi_H$, the other robots take actions using the robot policy $\pi_R$ trained in phase one.
The user can monitor the states of all robots simultaneously, and freely choose which robot to control using their internal scoring function $\phi$.
We observe which robot $i_H^t$ the user chooses to control at each timestep, and use these observations to infer the user's scoring function $\phi$.
In particular, we compute a maximum-likelihood estimate by fitting a parametric model $\hat{\phi} = \phi_{\bm{\theta}}$ that minimizes the negative log-likelihood loss function,
\begin{equation} \label{eq:mle}
\ell(\bm{\theta}; \mathcal{D}) = \sum_{(s_1^t, ..., s_n^t, i_H^t) \in \mathcal{D}} -\phi_{\bm{\theta}}\left(s_{i_H^t}\right) + \log{\left(\sum_{j=1}^n e^{\phi_{\bm{\theta}}\left(s_j^t\right)}\right)},
\end{equation}
where $\mathcal{D}$ is the training set of observed choices, and $\bm{\theta}$ are the weights of a feedforward neural network $\phi_{\bm{\theta}}$.\footnote{In our experiments, we used a multi-layer perceptron with two layers containing 32 hidden units each and ReLU activations.}
The learned scoring model $\hat{\phi}$ is optimized to explain the choices the user made in the training data, under the assumptions of the choice model in Equation \ref{eq:luce}.
Fitting a maximum-likelihood estimate $\hat{\phi}$ is the natural approach to learning a scoring model in our setting, since the MLE can be used to mimic the user's internal scoring function and thereby assist choice in phase three, and because it can be accomplished using standard supervised learning techniques for training neural networks.

\noindent\textbf{Phase three: assisted choice.}
At test time, we assist the user at time $t$ by automatically switching them to controlling the robot with the highest predicted likelihood of being chosen: robot $\argmax_{i \in [n]} \hat{\phi}(s_i^t)$.
This enables the user to manage a large number of robots $n$, where the user is unable to evaluate the scores of the states of all robots simultaneously using their internal scoring function $\phi$, but where we can trivially apply the learned scoring model $\hat{\phi}$ to the states of all robots simultaneously.

\noindent\textbf{Phase four (optional): improving the robot policy $\pi_R$.}
While performing the task in phase three, the user generates action demonstrations $(s, a)$ as they control each chosen robot -- demonstrations that can be added to the training data $\mathcal{D}_{\text{demo}}$ collected in phase one, and used to further improve the robot policy $\pi_R$ through online imitation learning.
One of the aims of assisted choice in phase three is to improve the quality of these additional demonstrations, since operator interventions in challenging states may provide more informative demonstrations.

\section{Simulation Experiments}

In our first experiment, we simulate human input, in order to understand how our method performs under ideal assumptions.
We seek to answer two questions: (1) does a model trained on data from a small fleet generalize to a large fleet, and (2) is our idea of treating choices as observations of preferences important for this generalization?
Simulating user input enables us to assess not just the task performance of our learned scoring model, but also its ability to recover the true internal scoring function; e.g., by posing counterfactual questions about how often the predictions made by our learned scoring model agree with the choices that would have been made by a simulated ground-truth scoring function.

\subsection{Experiment Design}

\noindent\textbf{Setup.}
We evaluate our method on a custom navigation task in the DOOM environment \cite{Kempka2016ViZDoom}.
In the navigation task, the robot navigates through a video game environment containing three linked rooms filled with hazards and health packs to reach a goal state (see screenshot in Figure \ref{fig:front} and schematic in Figure \ref{fig:study_perf}).
The robot receives low-dimensional observations $s \in \mathbb{R}^{4}$ encoding the robot's 2D position, angle, and health, and takes discrete actions that include moving forward or backward and turning left or right.
The default reward function outputs high reward for making progress toward the goal state and collecting health packs while avoiding hazards.
To introduce stochasticity into the environment, we randomize the initial state of each robot at the beginning of each episode.

\noindent\textbf{Simulating user input. }
Although simulating user input is not a part of our method, it is useful for experimentally evaluating our method under ideal assumptions.
We simulate the human operator with a synthetic user policy $\pi_H$ trained to maximize the environment's default reward function via deep reinforcement learning -- in particular, the soft actor-critic algorithm \cite{haarnoja2018sacapps}.
Note that our algorithm is not aware of the utility function $R$, and simply treats the reinforcement learning agent $\pi_H$ the same way it would treat a human user.
We choose the simulated ground-truth scoring function $\phi$ to be the gain in value from running the user policy $\pi_H$ instead of the robot policy $\pi_R$,
\begin{equation}
\phi(s_i^t) = V^{\pi_H}(s_i^t) - V^{\pi_R}(s_i^t),
\end{equation}
where $V$ denotes the value function, which we fit using temporal difference learning \cite{tesauro1995temporal} on the environment's default rewards.
Note that our choice of $\phi$ does not necessarily maximize the cumulative task performance of all robots, as assumed in Equation \ref{eq:cumulative-reward-objective} in Sec. \ref{problem-formulation}.
It is a heuristic that serves as a replacement for human behavior, for the purposes of testing whether we can learn a model of $\phi$ that performs as well as some ground-truth $\phi$.
We would like to emphasize that our method does not assume knowledge of $\pi_H$ or $\phi$, and that the design decisions made above are solely for the purpose of simulating user input in synthetic experiments.

\noindent\textbf{Manipulated variables.}
We manipulate the \emph{\textbf{scoring function}} used to select the robot for the synthetic user to control. The scoring function is either (1) the ground-truth scoring function $\phi$, (2) a model of the scoring function trained using our method $\hat{\phi}_{\text{luce}}$, or (3) a model of the scoring function trained using a baseline classification method $\hat{\phi}_{\text{base}}$.

Our method follows the procedure in Sec. \ref{our-method}: in phase two, it fits a scoring model $\hat{\phi}_{\text{luce}}$ to explain observations of the user's choices in a setting with a small number of robots $n = 4$, under the modeling assumptions in Sec. \ref{problem-formulation}.

The baseline method fits a scoring model $\hat{\phi}_{\text{base}}$ that assumes a much simpler user choice model than our method: that the user selects robot $i$ with probability $\sigma(\phi(s_i^t))$, where $\sigma$ is the sigmoid function.
In other words, the baseline method trains a binary classifier to distinguish between states where the user intervened and states where the user did not intervene.
Unlike our method, this approach does not model the fact that the user chose where to intervene based on relative differences in scores, rather than the absolute score of each robot.
Because of this modeling assumption, the baseline may incorrectly infer that a state $s_i^t$ was not worth intervening in because the user did not select robot $i$ at time $t$, when, in fact, the user would have liked to intervene in robot $i$ at time $t$ if possible, but ended up selecting another robot $j$ that required the user's attention even more than robot $i$.

We test each scoring function in a phase-three setting with a large number of robots $n = 12$.
At each timestep, we use the scoring function to choose which robot to control: the chosen robot executes an action sampled from the user policy $\pi_H$, while all other robots execute actions sampled from the robot policy $\pi_R$.

We also test each scoring function in a phase-four setting, where we re-train the robot's imitation policy $\pi_R$ using the newly-acquired user demonstrations from phase three, then evaluate $\pi_R$ by running it on a single autonomous robot.

\noindent\textbf{Dependent measures.}
To measure the \emph{\textbf{performance of the human-robot team}} in the phase-three setting with $n = 12$ robots, we compute the cumulative reward across all robots.
To measure the \emph{\textbf{predictive accuracy}} of each learned scoring model, we compute the top-1 accuracy of the robot ranking generated by the learned scoring models $\hat{\phi}_{\text{luce}}$ and $\hat{\phi}_{\text{base}}$ relative to the ranking produced by the ground-truth scoring function $\phi$.
To measure the \emph{\textbf{data impact}} of each scoring model on the quality of demonstrations used to improve the robot policy $\pi_R$ in phase four, we evaluate the task performance of the robot's imitation policy $\pi_R$ after being re-trained on the user action demonstrations generated in phase three.

\noindent\textbf{Hypothesis H1 (generalization)}. Our learned scoring model $\hat{\phi}_{\text{luce}}$ performs nearly as well as the ground-truth scoring function $\phi$ in the phase-three setting with a large number of robots, in terms of both the cumulative reward of the human-robot team and the data impact on the performance of a single robot.

\noindent\textbf{Hypothesis H2 (modeling relative vs. absolute preferences).} Our learned scoring model $\hat{\phi}_{\text{luce}}$ outperforms the baseline scoring model $\hat{\phi}_{\text{base}}$, in terms of all dependent measures -- cumulative reward, predictive accuracy, and data impact.

\subsection{Analysis.}

\begin{figure}[t]
\includegraphics[height=0.33\linewidth]{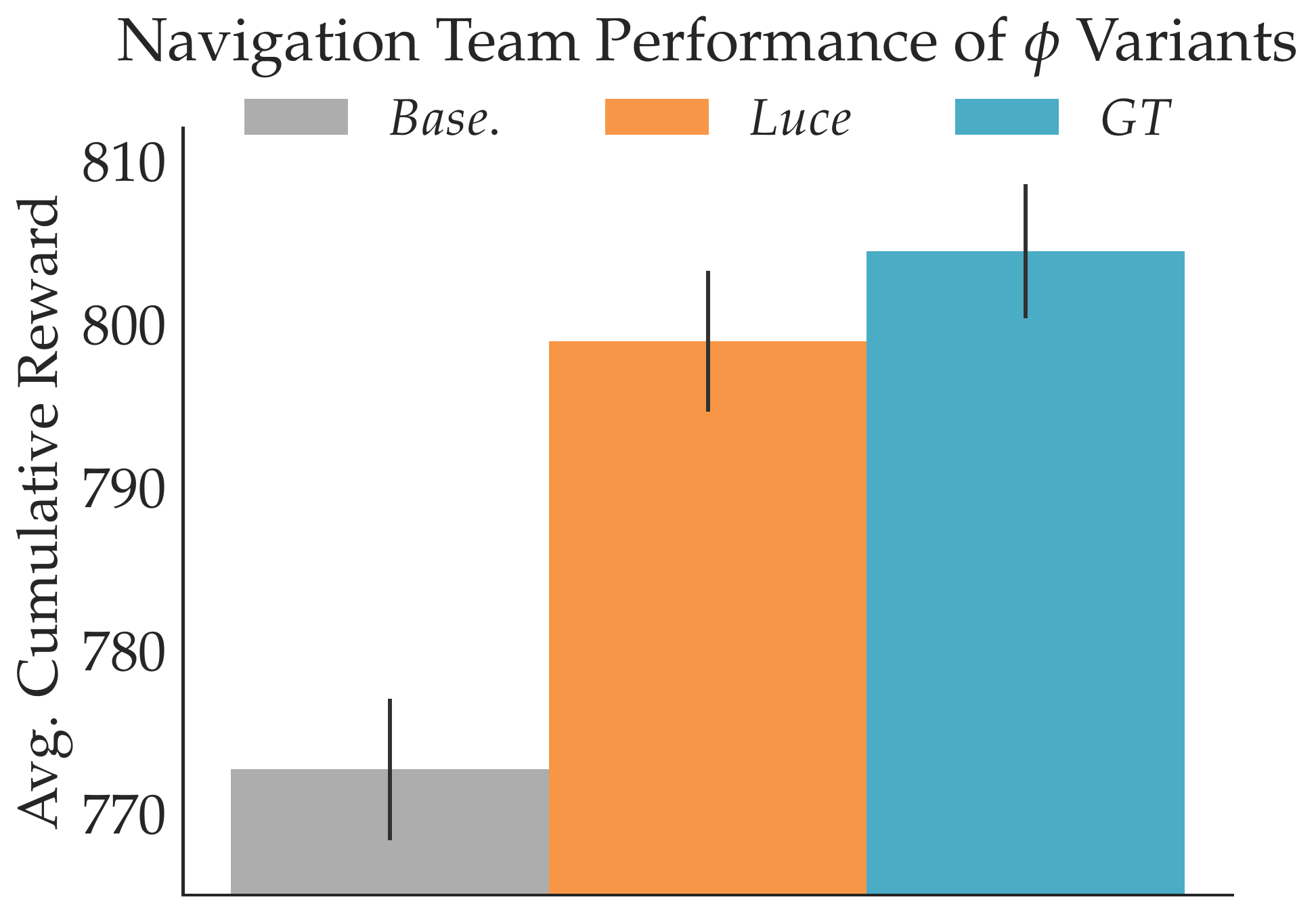}
  \includegraphics[height=0.33\linewidth]{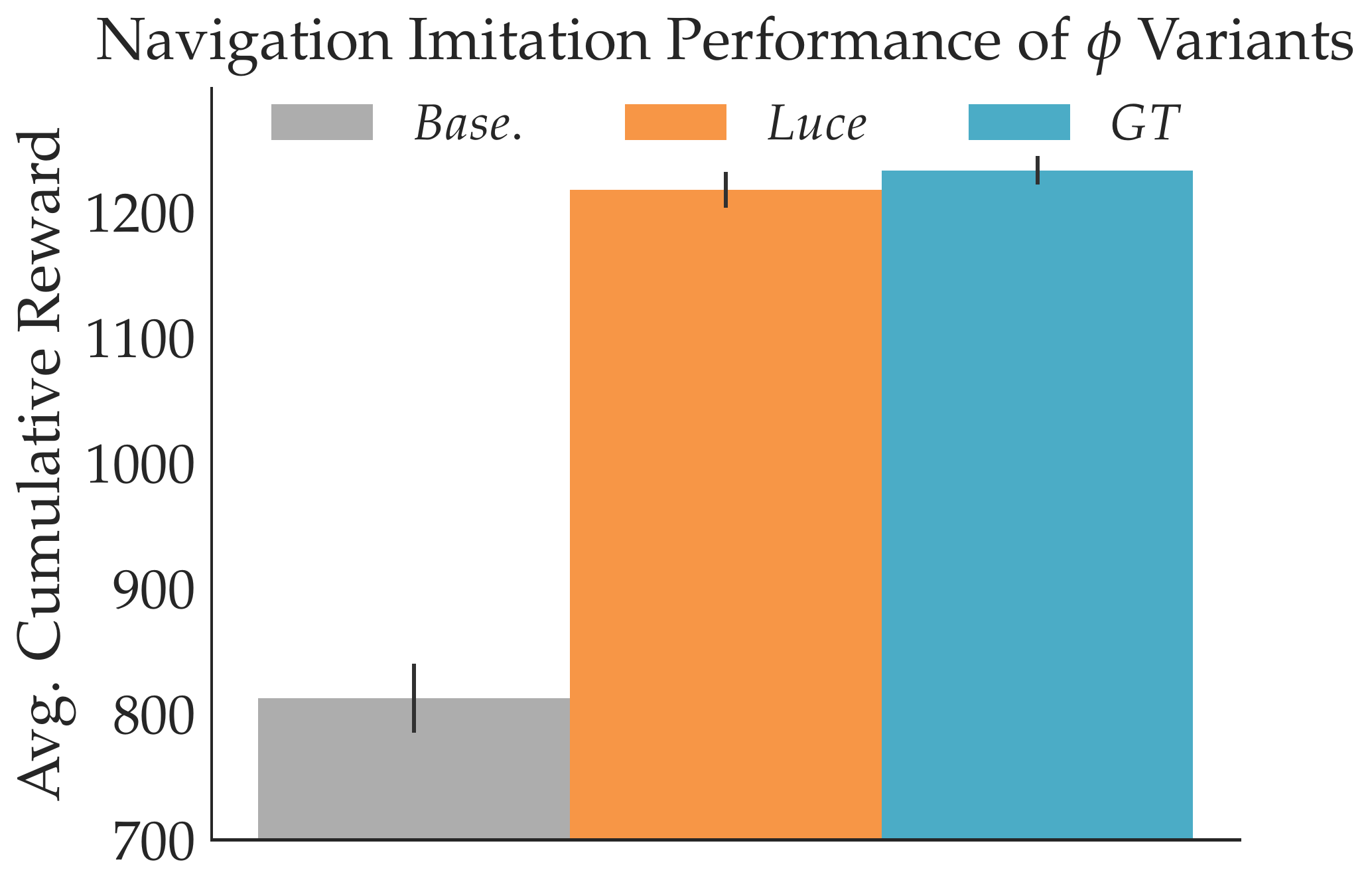}
  \caption{\textbf{Left}: our learned scoring model $\hat{\phi}_{\text{luce}}$ outperforms the baseline scoring model $\hat{\phi}_{\text{base}}$, while performing slightly worse than the ground-truth (GT) scoring function $\phi$. Rewards are averaged across all twelve robots, and across 250 trials. \textbf{Right}: the scoring function affects which states the user demonstrates actions in during phase three, which in turn affect the performance of the imitation policy after re-training with the new demonstrations in phase four. The ground-truth (GT) scoring function $\phi$ leads to the best imitation performance, followed by our method $\hat{\phi}_{\text{luce}}$. The baseline $\hat{\phi}_{\text{base}}$ induces less informative demonstrations. Rewards are for a single robot policy $\pi_R$ that runs without human intervention, and are averaged across 250 trials of 8 episodes each.}
  \label{fig:synth_imi_perf}
\end{figure}

Figure \ref{fig:synth_imi_perf} plot the performance and data impact of each scoring function for the navigation task.
In line with our hypotheses, $\hat{\phi}_{\text{luce}}$ outperforms $\hat{\phi}_{\text{base}}$ in all measures, while performing slightly worse than $\phi$.
In terms of predictive accuracy, we find that $\hat{\phi}_{\text{luce}}$ generalizes reasonably well: it agrees 79\% of the time with the ground truth, compared to 32\% for $\hat{\phi}_{\text{base}}$, which translates to better team performance.
Furthermore, demonstration data from assisted choice with $\hat{\phi}_{\text{luce}}$ induces a stronger imitation policy $\pi_R$. One explanation for this result is that collecting expert action demonstrations in challenging states leads to a better imitation policy $\pi_R$ than demonstrations in less challenging states.
Users tend to prefer to take over robots in states that are challenging for the robot policy $\pi_R$.
Our learned scoring model may capture that preference and, through assisted choice, allocate the user's expert actions to more challenging states.
These results suggest that our assisted choice method might be useful for active learning \cite{settles2009active} in the context of training robots via imitation \cite{argall2009survey} -- one possible direction for future work.

\section{User Study} \label{user-study}

The previous section analyzed our method's performance with synthetic data. Here, we investigate to what extent those results generalize to real user data.

\begin{figure}
\centering
\small
\begin{tabular}{lllll}
  & \textbf{Unassisted} & \textbf{Assisted} & F(1,11) & $p$-value \\
 \hline
 Q1 & 2.92 & 4.50 & 17.49 & $<\textbf{0.01}$ \\
 Q2 & 2.25 & 3.92 & 13.75 & $<\textbf{0.01}$  \\
 Q1$^\ast$ & 2.67 & 4.17 & 17.47 & $<\textbf{0.01}$  \\
 Q2$^\ast$ & 2.92 & 3.25 & 0.88 & 0.37 \\
\end{tabular}
\caption{Q1: it was easy to guide the robots to their goals. Q2: I was successful at guiding the robots. Q$^\ast$: Q after objective measures revealed. Survey responses on a 7-point Likert scale, where 1 = Strongly Disagree, 4 = Neither Disagree nor Agree, and 7 = Strongly Agree.}
\normalsize
\label{fig:likert}
\end{figure}

\begin{figure}
\centering
  \includegraphics[width=\linewidth]{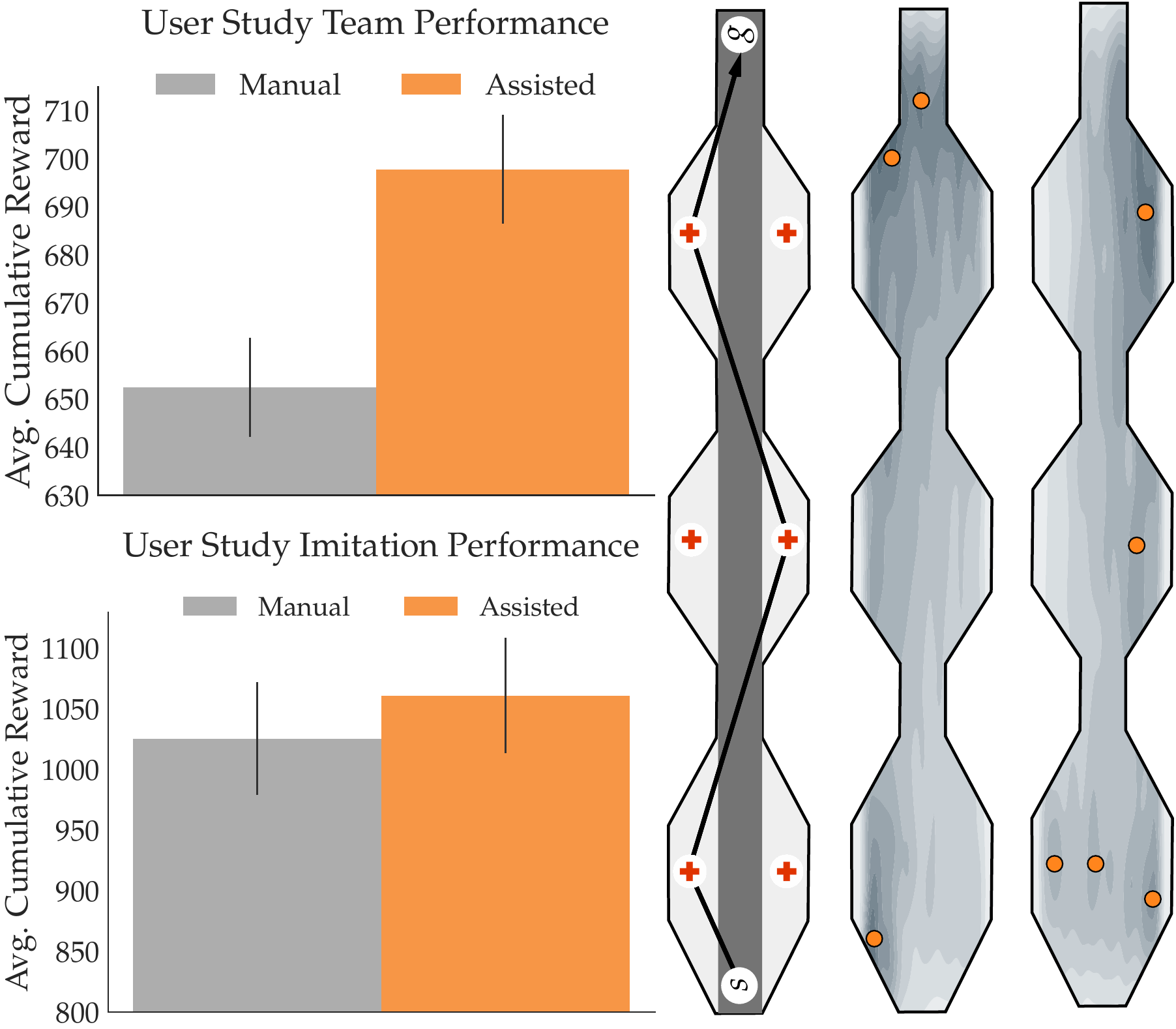}
  \caption{Performance of the human-robot team in phase three, averaged across twelve participants and eight trials per participant (top), and the robot's imitation policy in phase four (bottom). In the navigation task, the agent starts at $s$ and navigates to $g$, while avoiding the gray hazard region and collecting health packs indicated by crosses. The heat maps render the learned scoring models $\hat{\phi}$ of three different users, showing the positions where the model predicts each user would most prefer to take control of the robot. Darker indicates higher predicted score, and orange circles are peaks.}
  \label{fig:study_perf}
\end{figure}

\subsection{Experiment Design} \label{user-study-design}

\noindent\textbf{Setup.}
We use the navigation task for the study, which is split into the four phases described in Sec. \ref{our-method}.
In phase one, we collect data for training our robot policy $\pi_R$ via imitation: participants control a single robot -- initialized with a uniform random policy -- using the arrow keys on a keyboard, correcting it whenever it performs an action they would prefer it did not.
We do this for ten episodes.
In phase two, we collect data to train our scoring model $\hat{\phi}$: participants manage $n = 4$ robots, which they monitor simultaneously.
Participants manually choose which robot to control.
In phase three, participants manage $n = 12$ robots, and either manually choose which robot to control, or let the learned scoring model choose which robot to control for them.
In phase four, we re-train the imitation policy $\pi_R$ on the action demonstrations from phase three, and evaluate the task performance of $\pi_R$ in isolation.

\noindent\textbf{Manipulated variables.}
We manipulate whether or not the user is assisted by the learned scoring model $\hat{\phi}$ in choosing which of the $n = 12$ robots to control at any given time in phase three.
In the manual condition, the user is able to view the states of all robots simultaneously, and freely selects which robot to control using their internal scoring function $\phi$.
In the assisted condition, every fifteen timesteps the user is automatically switched to controlling the robot with the highest predicted likelihood of being chosen according to the learned scoring model $\hat{\phi}$.\footnote{We also tried a more flexible interface where, instead of automatically switching the user to the predicted robot, the interface continued showing all robots' states and merely highlighted the robot with the highest predicted likelihood of being chosen. The users in this pilot study tended to get confused by the suggestion interface, and preferred to be automatically switched to the predicted robot.}

\noindent\textbf{Dependent measures.}
As in the synthetic experiments, we measure \emph{\textbf{performance of the human-robot team}} using the cumulative reward across all robots, and \emph{\textbf{data impact}} using the reward achieved by a single robot running the imitation policy $\pi_R$ after training on the user demonstrations generated in the phase-three setting with $n = 12$ tobots.
We also conduct a survey using Likert-scale questions to measure subjective factors in the user experience, like the user's self-reported ease of use and perception of success with vs. without assisted choice.
We administered this survey after each condition, once before and once after revealing the cumulative reward to the participants.

\noindent\textbf{Hypothesis H3.}
We hypothesize that assisted choice will improve our objective and subjective dependent measures.

\noindent\textbf{Subject allocation.}
We conducted the user study with twelve participants, nine male and three female, with a mean age of twenty-one.
We used a within-subjects design and counterbalanced the order of the two conditions: manual choice, and assisted choice.

\subsection{Analysis.}
\noindent\textbf{Objective measures.}
We ran a repeated-measures ANOVA with the use of assisted choice (vs. manual choice) as a factor and trial number as a covariate on the performance of the human-robot team. Assisted choice significantly outperformed manual choice ($F(1,184)=12.96$, $p<.001$), supporting \textbf{H3} (see left plot in Figure \ref{fig:study_perf}).
Assisted choice also slightly improved the performance of the robot's imitation policy $\pi_R$ in phase four (see the right plot in Figure \ref{fig:study_perf}), although the difference was not statistically significant (F(1,23)=.31, p=.59).
One explanation for this result is that switching the user to a different robot every 15 timesteps was not frequent enough to produce the number of informative demonstrations needed to significantly improve $\pi_R$.

Figure \ref{fig:study_perf} illustrates the learned scoring models $\hat{\phi}$ of two different users in the study.
Each model is qualitatively different, showing that our method is capable of learning personalized choice strategies.
The model for the first user predicts they prefer to take control of robots near the goal position -- a strategy that makes sense for this particular user, since the autonomous robot policy trained in phase one was not capable of completing the task near the end.
The models for the second user predicts they would prefer to intervene near health packs, which makes sense since the robot policies trained to imitate that user tended to fail at maneuvering close to health packs.

\noindent\textbf{Subjective measures.}
Table \ref{fig:likert} shows that users self-reported that they found it easier to guide the robots to their goals and were more successful at guiding the robots in the assisted choice condition compared to the manual choice condition. The table also shows the results of running a repeated-measures ANOVA on each response. The improvement with our method was significant in all but one case: how successful users felt after we revealed their score to them.

\section{Hardware Demonstration}

\begin{figure}
\centering
  \includegraphics[height=0.43\linewidth]{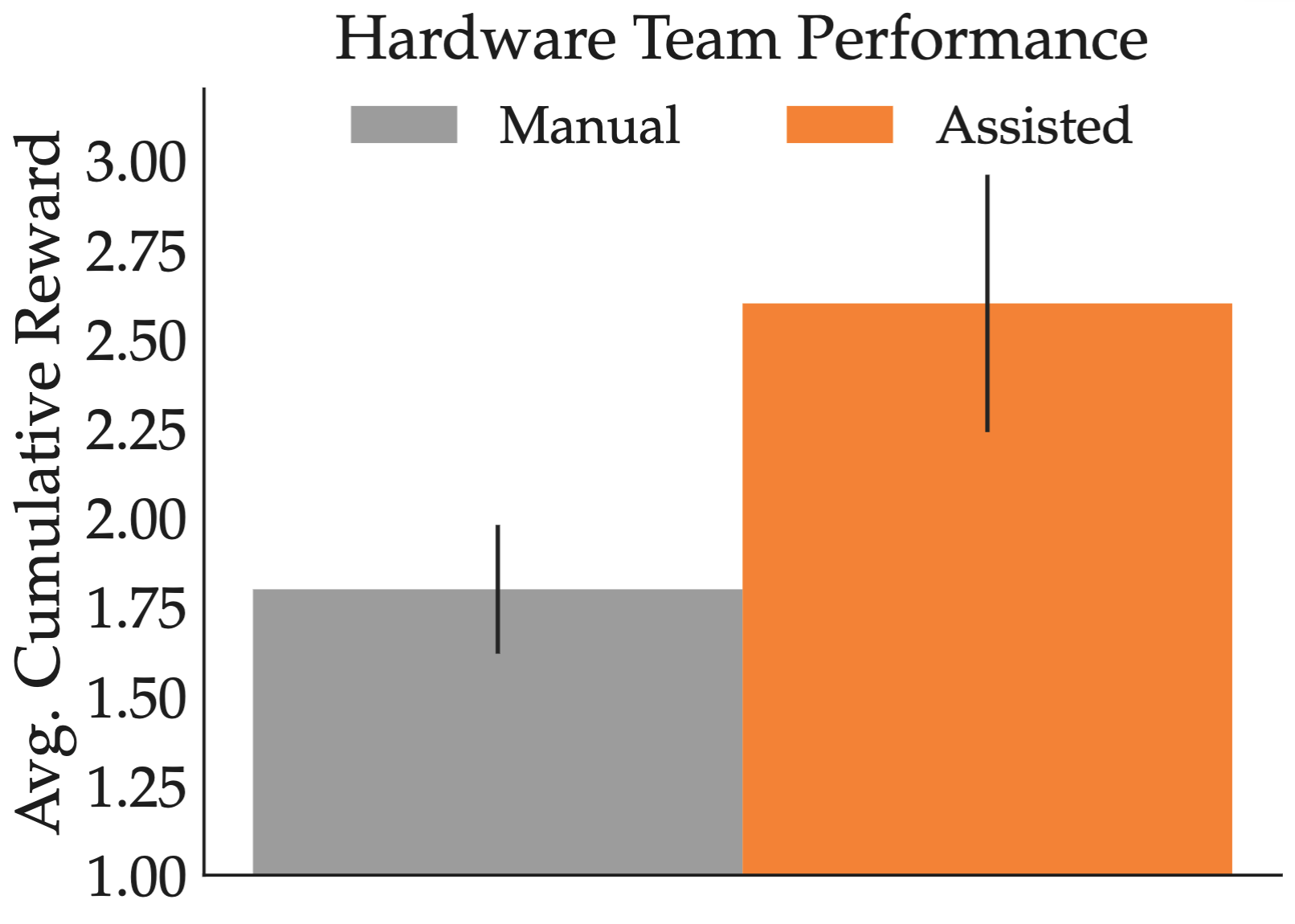}
  \includegraphics[height=0.43\linewidth]{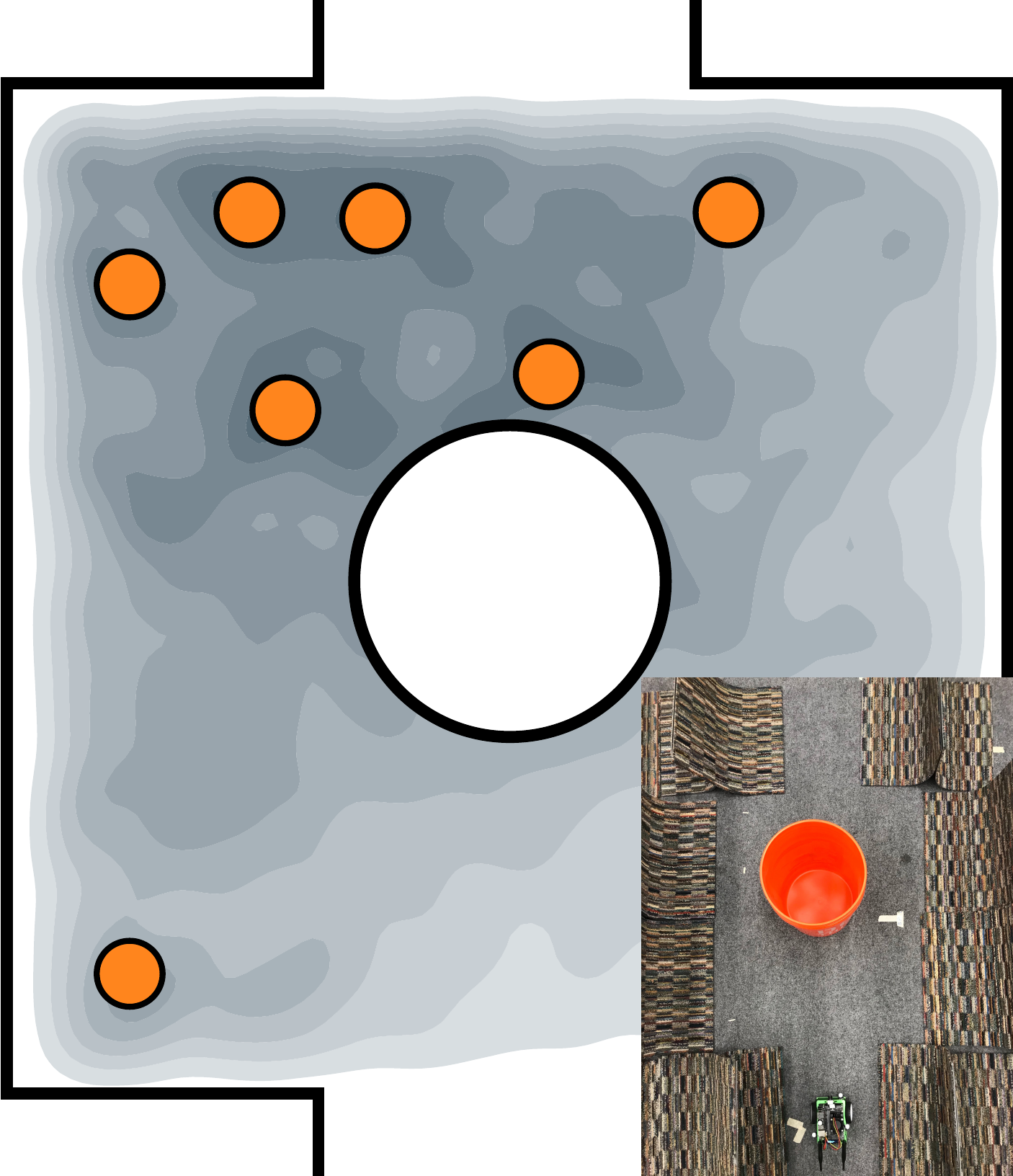}
  \caption{\textbf{Left}: performance of the human-robot team in phase three, averaged across five one-minute trials. \textbf{Right}: as in Figure \ref{fig:study_perf}, the heat map renders the learned scoring model $\hat{\phi}$.}
  \label{fig:hardware_results}
\end{figure}

The previous sections analyzed our method's performance on a simulated navigation task.
Here, we investigate to what extent those results generalize to real hardware.

We evaluate our method with a single human user driving laps with up to three NVIDIA JetBot mobile robots while avoiding centrally-placed buckets and surrounding obstacles (see Figures \ref{fig:front} and \ref{fig:hardware_results}).
We use an OptiTrack motion capture system to measure the positions and orientations of the robots.
We enable the user to control each robot's velocity -- via discrete keyboard actions for driving left, right, forward, and backward -- while looking through the robot's front-facing camera.
The reward function outputs $+1$ upon each lap completion.
Each episode lasts one minute.

We repeat the three phases from Sec. \ref{user-study}.
In phase one, we collect 2000 state-action demonstrations for training our robot policy $\pi_R$ via imitation.
In phase two, we collect 1300 examples of a human's preferences $(s_1^t, s_2^t, i_H^t)$ to train our scoring model $\hat{\phi}$: the human -- an author, for the purposes of our demonstration -- manages $n = 2$ robots, which they monitor simultaneously.
In phase three, they manage $n = 3$ robots, and either manually choose which robot to control, or let the learned scoring model choose which robot to control for them every five seconds.

As in Sec. \ref{user-study}, we manipulate whether or not the user is assisted by the learned scoring model $\hat{\phi}$ in choosing which of the $n = 3$ robots to control at any given time in phase three.
We measure \emph{\textbf{performance of the human-robot team}} using the cumulative reward across all robots.

We find that again, assisted choice improves the performance of the human-robot team in phase three (Figure \ref{fig:hardware_results}).
The learned scoring model $\hat{\phi}$ helpfully predicts high scores for states close to the bucket, the boundary of the environment, or a corner where the robot tends to get stuck, where user interventions tend to be concentrated during phase two.

\section{Limitations and Future Work}

We have evaluated our method in a simulated environment, and demonstrated its feasibility on real hardware in a controlled lab setting.
For complex, real-world tasks, it may be the case that learning a scoring model is as difficult as learning the user's policy.
In such cases, exploiting our method's ability to gather useful demonstration data by focusing the user's attention on informative states would be an interesting area for further investigation.

Another direction for future work could be to investigate tasks where users tend to change their control policies and internal scoring models as the number of robots increases, and whether iteratively applying our method addresses non-stationary user models.
\addtolength{\textheight}{-12cm}   

\section*{ACKNOWLEDGMENT}
This work was supported in part by NSF SCHOOL, Intel, Berkeley DeepDrive, GPU donations from NVIDIA, NSF IIS-1700696, AFOSR FA9550-17-1-0308, NSF NRI 1734633, an NVIDIA Graduate Fellowship. The authors would like to thank Justin Yim for his help in setting up motion capture for the hardware experiment.

\bibliographystyle{IEEEtran}
\bibliography{IEEEabrv,IEEEexample}
\end{document}